%% file: main.tex
\documentclass[10pt,twocolumn,letterpaper]{article}

\usepackage{wacv}
\usepackage{times}
\usepackage{epsfig}
\usepackage{graphicx}
\usepackage{amsmath}
\usepackage{amssymb}
\usepackage[utf8]{inputenc} 
\usepackage[T1]{fontenc}    
\usepackage{url}            
\usepackage{booktabs}       
\usepackage{amsfonts}       
\usepackage{amsmath}
\usepackage{amssymb}
\usepackage{nicefrac}       
\usepackage{microtype}      
\usepackage[pagebackref=true,breaklinks=true,letterpaper=true,colorlinks,bookmarks=false]{hyperref}
\usepackage{lib}
\usepackage{overpic}
\usepackage{graphicx}
\usepackage{subfig}
\usepackage{verbatim}
\usepackage[font=footnotesize,labelfont=bf]{caption} 

\newcommand{\denselist}{\itemsep 0pt\parsep=0pt\partopsep 0pt\vspace{-\topsep}}
\newcommand{\bitem}{\begin{itemize}\denselist}
\newcommand{\eitem}{\end{itemize}}

\newcommand\blfootnote[1]{%
  \begingroup
  \renewcommand\thefootnote{}\footnote{#1}%
  \addtocounter{footnote}{-1}%
  \endgroup
}

\usepackage[pagebackref=true,breaklinks=true,letterpaper=true,colorlinks,bookmarks=false]{hyperref}

\def\wacvPaperID{168} 

\wacvfinalcopy 
\begin{document}

\title{Object-Centric Multi-View Aggregation}

\author{
Shubham Tulsiani$^{1}$ \qquad 
Or Litany$^{2\dag}$ \qquad 
Charles R. Qi$^{\dag}$\qquad 
He Wang$^{2\dag}$\qquad Leonidas J. Guibas$^{2\dag}$\\
$^1$Facebook AI\qquad $^2$Stanford University}

\maketitle

\begin{abstract}
We present an approach for aggregating a sparse set of views of an object in order to compute a semi-implicit 3D representation in the form of a volumetric feature grid. Key to our approach is an object-centric canonical 3D coordinate system into which views can be lifted, without explicit camera pose estimation, and then combined -- in a manner that can accommodate a variable number of views and is view order independent. We show that computing a symmetry-aware mapping from pixels to the canonical coordinate system allows us to better propagate information to unseen regions, as well as to robustly overcome pose ambiguities during inference. Our aggregate representation enables us to perform 3D inference tasks like volumetric reconstruction and novel view synthesis, and we use these tasks to demonstrate the benefits of our aggregation approach as compared to implicit or camera-centric alternatives.
\end{abstract}

\blfootnote{\dag: work done while at Facebook.}

\input{intro_aggregation}

\input{related}

\input{approach}

\input{experiments}

\input{discussion}

{\small
\bibliographystyle{ieee_fullname}
\bibliography{ref}
}

\end{document}


\title{Supplementary: Object-Centric Multi-View Aggregation}

\author{First Author\\
Institution1\\
Institution1 address\\
{\tt\small firstauthor@i1.org}
\and
Second Author\\
Institution2\\
First line of institution2 address\\
{\tt\small secondauthor@i2.org}
}

\maketitle

\input{appendix}

{\small
\bibliographystyle{ieee}
\bibliography{ref}
}

%% file: intro_aggregation.tex
\section{Introduction}
A central problem in computer vision is to recover the 3D structure of the world from 2D views of it, namely images. Classical approaches such as Structure from Motion (SfM) operate in the setting where many views (say tens to hundreds) are available so that geometric correspondences between them can be used to infer camera pose and induce 3D structure~\cite{tomasi1992shape,koenderink1991affine}. More recently there has been a flurry of activity on object reconstruction from a single image, exploiting machine learning and especially deep learning approaches to hallucinate information invisible in that view~\cite{girdhar16b,choy20163d,fan2017point,tulsiani2017multi}, building on large 3D model repositories such as ShapeNet~\cite{chang2015shapenet} for training. We aim to tackle the scenario in between these two extremes, and addresses situations where a few images (say one to four) are available \eg, online product marketplaces.

In this work, we focus on the single-object setup: how to recover a `3D representation' of the underlying object given one or more images. The key questions for this task pertain to the form of this 3D representation, and how can one enable aggregation of the information across views to compute a unified representation. While classical methods pursue explicit representations such as point clouds and aggregate views via explicit correspondence inference, these choices are not easily applicable to our setup with a small number of input images with unknown camera poses. In contrast, learning based methods typically represent 3D implicitly~\cite{girdhar16b,choy20163d} \eg via a single latent vector, and can be extended to implicitly aggregate images \eg, via an LSTM~\cite{choy20163d}. However, this fully implicit representation and aggregation ignores the underlying geometric structure of the task. We instead pursue semi-implicit 3D representations~\cite{sitzmanndeepvoxels}, as these combine a 3D voxel grid with an implicit latent vector in each voxel coding the contents of that cell, and allow easily recovering explicit structure. We propose an aggregation mechanism to infer such a semi-latent representation given multiple images, and show that it allows us to perform 3D centric tasks \eg shape inference or novel view synthesis.

\begin{figure*}[t]
    \centering
    \includegraphics[width=1\linewidth]{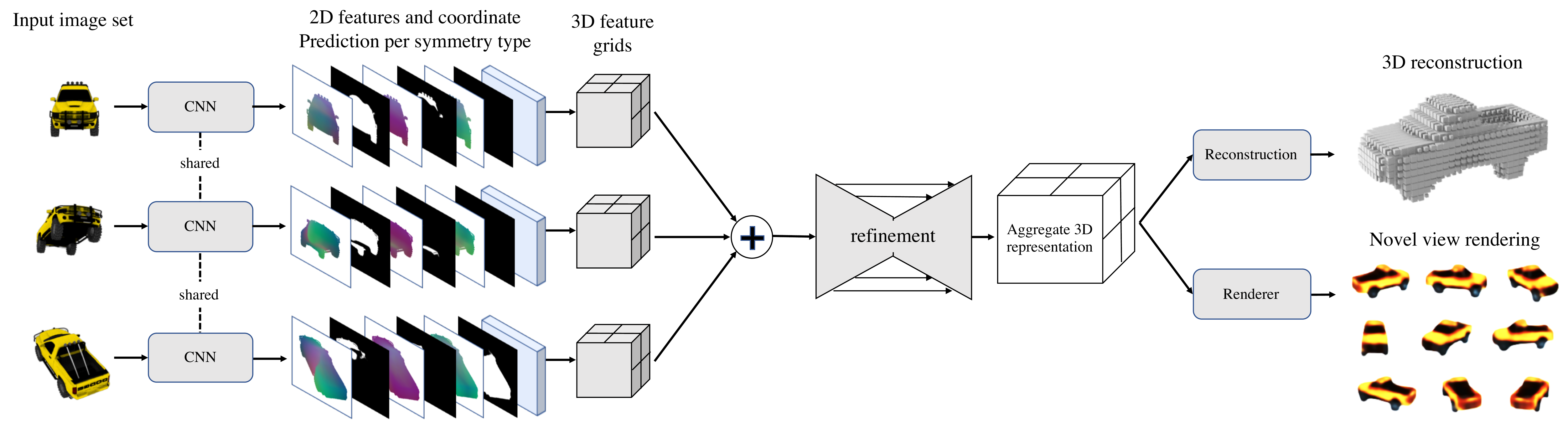}
    \caption{\label{fig:mva_teaser} \textbf{Multi-view aggregation and its applications.} An input image set is first passed through a 2D CNN that outputs per-pixel features, and for each symmetry type a dense (pre-pixel) coordinate prediction (3-dim) and a confidence map (1-dim). Three pairs of predicted coordinates and confidence maps are shown as an illustration. The computed features are then lifted to 3D voxel locations prescribed by the coordinate prediction and weighted by the confidences. A 3D-UNet is then used to refine the averaged 3D features to yield an aggregate 3D representation, which is further used for (a) 3D reconstruction, and (b) novel-view synthesis.}
\end{figure*}

Our key insight towards designing the aggregation mechanism is that for each pixel, we can `lift' it to a canonical object-centric space, and then process the information from across the images in this shared canonical space. Unlike more traditional computer vision techniques that are camera-focused and need to know or estimate the camera pose for each view in a world coordinate system so as to properly integrate them, we instead accomplish view aggregation without explicit camera pose estimation, by directly lifting views into an object-centric space. This further allows exploitation of prior object knowledge that is difficult to incorporate in camera-centric approaches. For example, many objects possess important symmetries that can be inferred from knowledge of the object class and the observed view or views. With such knowledge, when we lift a given view into the canonical object-centric space in 3D, we can augment the observed regions by their inferred symmetric counterparts, effectively enhancing our understanding of the 3D structure of the object and allowing us to produce a complete 3D representation from fewer views. Critically, symmetry induced augmentation also solves the problem of pose ambiguities due to object symmetries or part occlusions (e.g., resolving `front' or `back' of a bottle) -- by effectively generating in the object coordinate space the union of the lift from all possible valid poses, thus making such ambiguities immaterial.

Our overall architecture consists of three parts: a) a network that lifts a single image into a semi-implicit representation of the seen object in a voxelized object-centric 3D space annotated with learned features per voxel, b) a symmetric view aggregator that can combine lifts from different views into a single semi-implicit representation, and c) task specific networks that then use this representation for different downstream tasks, such as volumetric reconstruction or novel view synthesis.

In summary, we present an approach to infer a 3D representation given a sparse set of images with unknown camera poses, and show that this representation can be used for downstream 3D tasks. Our key contributions are:
\bitem
    \item By using an object-centric coordinate system we obviate the need for camera pose estimation in 3D inference and view aggregation.
    \item Our symmetry-aware view lifting into the object-centric space allows us to extract more information from each view, while also bypassing ambiguities.
    \item Our semi-implicit 3D representation carries a strong geometric inductive bias in its formulation, and combines latent local elements that that can be trained in an end-to-end fashion for multiple downstream tasks. 
\eitem

%% file: related.tex
\section{Related Work}
The task of recovering the underlying 3D structure from multiple images is a classical one in the computer vision community. Early approaches can broadly be categorized as tackling Multi-view Stereopsis (MVS)~\cite{seitz2006comparison,furukawa2015multi}, where the aim is to recover 3D given camera poses, or Structure from Motion (SfM)~\cite{ullman1979interpretation,haming2010structure}, where the camera pose is also unknown. A long line of work in both setups~\cite{brown2005unsupervised,snavely2006photo,furukawa2009accurate,schonberger2016pixelwise}, relying on cues such as geometric constraints and photometric consistency, has yielded impressive results, but in scenarios where numerous views of a scene/object are available. However, due to the reliance on such cues, these learning-free methods are not well-suited for handling a sparse set of images with possibly little overlap. We instead aim to build a system that can retain the geometric inductive biases of these methods, while also leveraging data-driven priors via learning to tackle inference from sparse set of images with unknown camera poses.

A scenario where these data-driven priors have been successfully exploited by learning based approaches is that of 3D prediction from a single input image. Driven by the success of deep learning, several methods tackle the task of inferring volumetric~\cite{choy20163d,girdhar16b,tulsiani2017multi,marrnet,hsp}, mesh~\cite{kanazawa2018learning,pontes2017image2mesh,deepMarching,wang2018pixel2mesh}, point cloud~\cite{fan2017point,lin2018learning}, or implicit 3D~\cite{groueix2018papier,mescheder2018occupancy} representations from an input image. These approaches learn deep neural network based models that during inference, directly predict the underlying 3D in a feed forward manner given a single image, without explicitly relying on the geometry of the task. Our goal in this work is to extend these approaches, in particular ones for volumetric inference, to leverage multiple input images, and we do so using a geometrically motivated aggregation mechanism.

There have been several recent attempts~\cite{lsmKarHM2017,huang2018deepmvs,yao2018mvsnet,paschalidou2018raynet,saito2019pifu,huang2020arch,olszewski2019transformable,yariv2020multiview} which, sharing our motivation for integrating geometric inductive biases in a learning framework, tackle the task of learning based multi-view stereopsis. While these demonstrate impressive results, they all crucially rely on the availability of the ground-truth camera poses during inference. We instead pursue the task of 3D reconstruction with only a set of images at inference, without the availability of known camera poses. In contrast to these approaches which rely on global camera estimates and camera-centric predictions \eg depth, we propose to directly predict object-centric coordinates. Closer to our setup is the work by Choy \etal~\cite{choy20163d} which similarly infers 3D from multiple images without known camera poses, but performs implicit feature aggregation, whereas we propose a more geometrically motivated mechanism. More recently, work from Sridhar et. al. \cite{sridhar2019multiview} also investigated the task of aggregating multiple views, but in contrast to our approach, did not account for the ambiguities due to symmetry.

%% file: approach.tex
\section{Approach}
Given multiple images of an object with associated foreground masks, we aim to compute an aggregate representation that incorporates information from across the images, and then leverage this representation for tasks such as 3D reconstruction and novel view synthesis. Our key insight is that we can first `lift' the information from the foreground pixels across different images of an instance to an object-centric 3D space, and then aggregate and process the features from across the images in this shared canonical space. This yields a canonicalized 
3D representation for the underlying object, which can then be used towards various 3D related tasks.

Our proposed system, as depicted in Figure \ref{fig:mva_teaser}, is comprised of three stages: a) lifting pixels to an object-centric 3D space with symmetry augmentation, b) multi-image feature aggregation, and c) task-specific computation. We first present the lifting procedure in \secref{lifting} where we learn a (probabilistic) mapping from pixels to coordinates in a normalized space, while allowing symmetries to overcome ambiguities. We then describe in \secref{aggregation} how these learned mappings enable copying pixel-wise features from across the images into a volumetric feature grid, where these are aggregated and refined via a learned function. Finally, we show in \secref{tasks3d} that the aggregate object representation can be leveraged for various 3D-centric tasks.

\subsection{Symmetry-Aware Object-Centric Coordinate Prediction}
\seclabel{lifting}
We aim to compute a mapping from pixels to a canonical 3D space to enable aggregation across images. We build on the insight by Wang \etal~\cite{wang2019normalized} that such a mapping can be defined using a normalized and aligned shape collection. We briefly summarize the canonical coordinate prediction ~\cite{wang2019normalized} into an object-centric space, and describe in detail our symmetry-aware formulation that allows overcoming ambiguities in the task, and the corresponding training objective.

\paragraph{Object-Centric Coordinate Prediction.}
Each foreground pixel corresponds to a 3D point on the underlying object surface. Using a normalized shape collection, where all shapes are aligned and scaled to fit unit diagonal cubes, one can learn a mapping from pixels to their corresponding coordinates in this space. Given an image $I$, we can learn to predict a pixel-wise mapping $C$, where for a pixel $u$, $C[u] \in \mathbb{R}^3$ corresponds to the canonical coordinate of the 3D point visible at that pixel. Using synthetically rendered images, it is easy to obtain the ground-truth mapping $\hat{C}$ and learn a parametric prediction function $f_\theta(I) \equiv C$ using such supervision. While Wang \etal~\cite{wang2019normalized} leveraged such a mapping for the task of pose estimation from a single image, we observe that it can also allow us to aggregate information across several images of an instance via lifting pixels (and associated features) into this canonical space.

\paragraph{Overcoming Ambiguities via Allowing Symmetries.}
Unfortunately, inferring this mapping from pixels to their canonical coordinates for generic objects is an inherently ambiguous task. Consider a pixel on the leg of a symmetrical square table with four legs. While this pixel does have a unique canonical coordinate (as defined by the the position of the corresponding 3D point in the normalized object model), it is not possible to infer this coordinate given an image alone. Such ambiguities are common due to local and global symmetries across objects.

Our insight is that instead of predicting a single canonical coordinate for each pixel, we can instead learn to predict a \textit{symmetry-aware distribution}. This allows us to both: a) overcome mean prediction effects when dealing with ambiguities, b) propagate information from a pixel to multiple 3D locations depending on the inferred symmetries. We therefore make predictions of coordinates and probabilities for multiple symmetry types. We assume a set of possible symmetry types $\mathcal{G}$, with each $g \in \mathcal{G}$ indicating either a rotational or reflection symmetry. We overload notation, and also use $g$ to denote a function that, given an input point $x \in \mathbb{R}^3$, generates the corresponding set of points under the symmetry type $g$, \ie $g(x)$ is the closure set for $x$ under $g$. Concretely, we consider 5 global symmetry types: identity, reflection along $y$ axis, and $2, 4, $ or continuous rotational symmetry along z-axis. The first type corresponds to no symmetry prediction, while the others correspond to some commonly occurring global symmetries across objects.


Given an input image $I$, we predict $f_\theta(I) \equiv \{(C^g, P^g)\}$, where for a pixel $u$, $P^g[u]$ denotes the probability that the underlying 3D point belongs to the symmetry type $g$ (the per-pixel probabilities sum to 1), and $C^g[u]$ indicates its object-centric canonical coordinate if it does. Note that this is equivalent to predicting the set of points $g(C^g[u])$ with probability $P^g[u]$ at the pixel $u$.

\paragraph{Training Objective.}
We can learn this per-symmetry type canonical coordinate prediction using supervision in the form of the ground-truth coordinates $\hat{C}$. Note that this is similar supervision as in the case of learning canonical prediction without allowing symmetries, hence \emph{we do not need to rely on explicit supervision for the per-type coordinate or probability predictions}. 
Our training objective comprises two terms that work together to (a) encourage the predicted symmetry type at each pixel to contain the correct coordinate, while (b) penalizing spurious predictions.

For a pixel $u$, we penalize the distance between its associated canonical coordinate $\hat{C}[u]$ and the closest point in each symmetry type predicted, and weight this loss by the corresponding probability.
\begin{gather}
\eqlabel{lcoord}
    L_{c} = \sum_{u} \sum_{g} P^g[u] ~ \min_{x \in g(C^g[u])} \|x - \hat{C}[u]\|\,.
\end{gather}
This loss enables overcoming ambiguities inherent in the task due to symmetries, as instead of requiring the prediction of the true canonical coordinate, it only requires predicting a possible canonical coordinate in the respective symmetry type.

However, this encourages over-predicting symmetry types as the additional points can only reduce the loss. We therefore introduce a second loss that reduces spurious predictions by penalizing them for inducing points that do not exist in the underlying 3D shape.  Denoting by $\mathcal{D}(S, x)$ the distance from a point $x$ to its closet point on a shape $S$, the additional objective is:
\begin{gather}
\eqlabel{lchamfer}
    L_{s} = \sum_{u} \sum_{g} P^g[u] ~ \max_{x \in g(C^g[u])} \mathcal{D}(S, x)\,.
\end{gather}
We use a UNet~\cite{Ronneberger_2015} based CNN as the parametrized predictor $f_{\theta}$ to learn the symmetry-aware canonical coordinate prediction. As an implementation detail, while it is easy to analytically compute the objective in \eqref{lcoord} for the symmetry types we consider, we use a finite number of random samples for symmetry types that induce sets with infinite cardinality for the objective in \eqref{lchamfer}.

\subsection{Multi-Image Feature Aggregation}
\seclabel{aggregation}
Given the learned (probabilistic) mapping from pixels across the images to a canonical 3D space, we can compute an aggregate representation in the form of a volumetric feature grid. We do so via first computing 2D per-pixel features across the input images and lift these features to a shared 3D grid using the inferred embeddings. We can then combine and further process the features from the different images in this volumetric space to obtain an aggregate representation that encompasses the information from across the input images. Concretely, given $K$ images $\{I_k\}$ of an instance, we first compute corresponding 2D features $\{F_k\}$. We then lift these features using the predicted canonical coordinates to a per-image volumetric feature grid $V_k$, and then aggregate these to obtain a volumetric feature representation $V$.

\paragraph{2D Feature Extraction.} We want a 2D encoder that can capture global context while preserving the low-level details. Towards this, we extend the UNet~\cite{Ronneberger_2015} $f_{\theta}$ presented in \secref{lifting} to additionally output per-pixel features $F$ given input image $I$.

\paragraph{Probabilistic Feature Splatting.} The predicted canonical coordinates associated with image $I_k$, $\{(C^g_k, P^g_k)\} \equiv f_\theta(I_k)$ allow us to lift the associated 2D features $F_k$ to a volumetric feature grid $V_k$. Intuitively, we start with an empty 3D feature grid and for each pixel, we add the associated 2D feature (weighted by the corresponding symmetry type probability) at the 3D location(s) implied by the canonical coordinate prediction. Note that a single pixel may lead to placing features at multiple 3D locations based on the underlying symmetry type.

We denote by $\mathcal{V}(x, f)$ a 3D feature grid obtained by placing a feature $f$ at the 3D coordinate $x$ in an initially empty grid. Note that this grid is empty at all locations except up to 8 cells immediately around the coordinate $x$ (see appendix for details). Using this notation, we can define our lifted feature grid as the probability weighted combination of all the 3D grids obtained for each pixel.
\begin{gather}
\eqlabel{featlift}
V_k = \sum_{u} \sum_{g} \sum_{x \in g(C^g_k[u])} P^g_k[u]~\mathcal{V}(x, F_k[u])\,.
\end{gather}

This procedure allows us to copy features from pixels to possibly multiple locations as implied by the symmetry predictions. Further, for use in aggregating feature grids across images, we also compute a `weight' grid that records the total number of pixels that contribute to each cell (weighted by probabilities).
\begin{gather}
\eqlabel{weightlift}
W_k = \sum_{u} \sum_{g} \sum_{x \in g(C^g_k[u])} P^g_k[u]~\mathcal{V}(x, 1)\,.
\end{gather}

\paragraph{Averaging and Refinement.}
Having obtained feature and weight grids $\{(V_k, W_k)\}$ for each input image $I_k$, we can now construct a sum weight and an average feature across the images:
\begin{gather}
\eqlabel{featavg}
\bar{W} = \sum_{k} W_k ~~~;~~~ \bar{V} = \frac{\sum_{k}V_k}{\bar{W}}\,,
\end{gather}
Where the division is understood as a voxel-wise operation. Finally, we use a 3D UNet based CNN $h_{\psi}$ to process the features and yield a final aggregate representation $V$ that incorporates information from all the views. This additional processing allows us to perform 3D reasoning using the lifted 2D features, and can implicitly perform noise filtering \etc and also propagate information to regions of the 3D volume without any direct image evidence.

\begin{gather}
\eqlabel{featprocess}
V = h_{\psi}([\bar{V}; \bar{W}])\,.
\end{gather}
\subsection{Learning 3D-Centric Tasks}
\seclabel{tasks3d}
Our aggregate feature representation $V$ is a volumetric feature grid that integrates information from multiple images of the object. We leverage this representation for learning 3D tasks, and show how we can train our system using tasks like volumetric prediction and novel view synthesis as supervision.

\paragraph{Volumetric 3D Prediction.}
A task pursued by previous multi-image prediction systems~\cite{choy20163d} is that of 3D reconstruction. We show that our aggregate representation can also be used for this task, and that our approach improves over fully implicit aggregation mechanisms. We use a lightweight 2-layer 3D CNN to predict voxel occupancy $O$ from our volumetric representation and use a cross-entropy loss $L_{vol}$ between the ground-truth and predicted occupancies.

\paragraph{Novel View Synthesis.}
To demonstrate that the aggregate representation can capture the \textit{appearance} of 3D objects, we synthesize novel views from it. 

We adopt the pipeline from ~\cite{sitzmanndeepvoxels} and train a renderer $\mathcal{R}$ that yields an image $\mathcal{R}(V, \pi)$ given an input feature grid $V$ and camera viewpoint $\pi$.

While ~\cite{sitzmanndeepvoxels} used this renderer in conjunction with an optimized feature grid that required hundreds of input views with known camera poses, we adopt it to our setting where the 3D representation is predicted from only few images with unknown camera poses. As supervision for training, we assume novel views of the object $\{I_{k'}, \pi_{k'}\}$ where the image $I_{k'}$ has a corresponding camera viewpoint $\pi_{k'}$. Our view synthesis loss is defined as: $L_{vs} = \sum_{k'} \|\mathcal{R}(V, \pi_{k'}) - I_{k'} \|\,.$ Note that while the novel views used for supervising the rendering have known camera viewpoints, the images used to compute the aggregate representation do not.
\paragraph{Overall Training Objective.}
Our training objective comprises terms for learning the canonical coordinate prediction $(L_{s}, L_{c})$ as well as task-specific reconstruction and rendering objectives $(L_{vol}, L_{vs})$. We weight the loss terms to (approximately) equalize their contribution to the total loss. Additionally, we find it beneficial to decouple the learning of the coordinate prediction from the downstream tasks, \ie the task-specific losses only influence the learned feature representations. We learn a common model across all categories, trained jointly for both tasks (reconstruction and view synthesis). We will publicly release our implementation for reproducibility.

%% file: experiments.tex
\begin{figure*}[t]
\center
\includegraphics[width=1.04\linewidth]{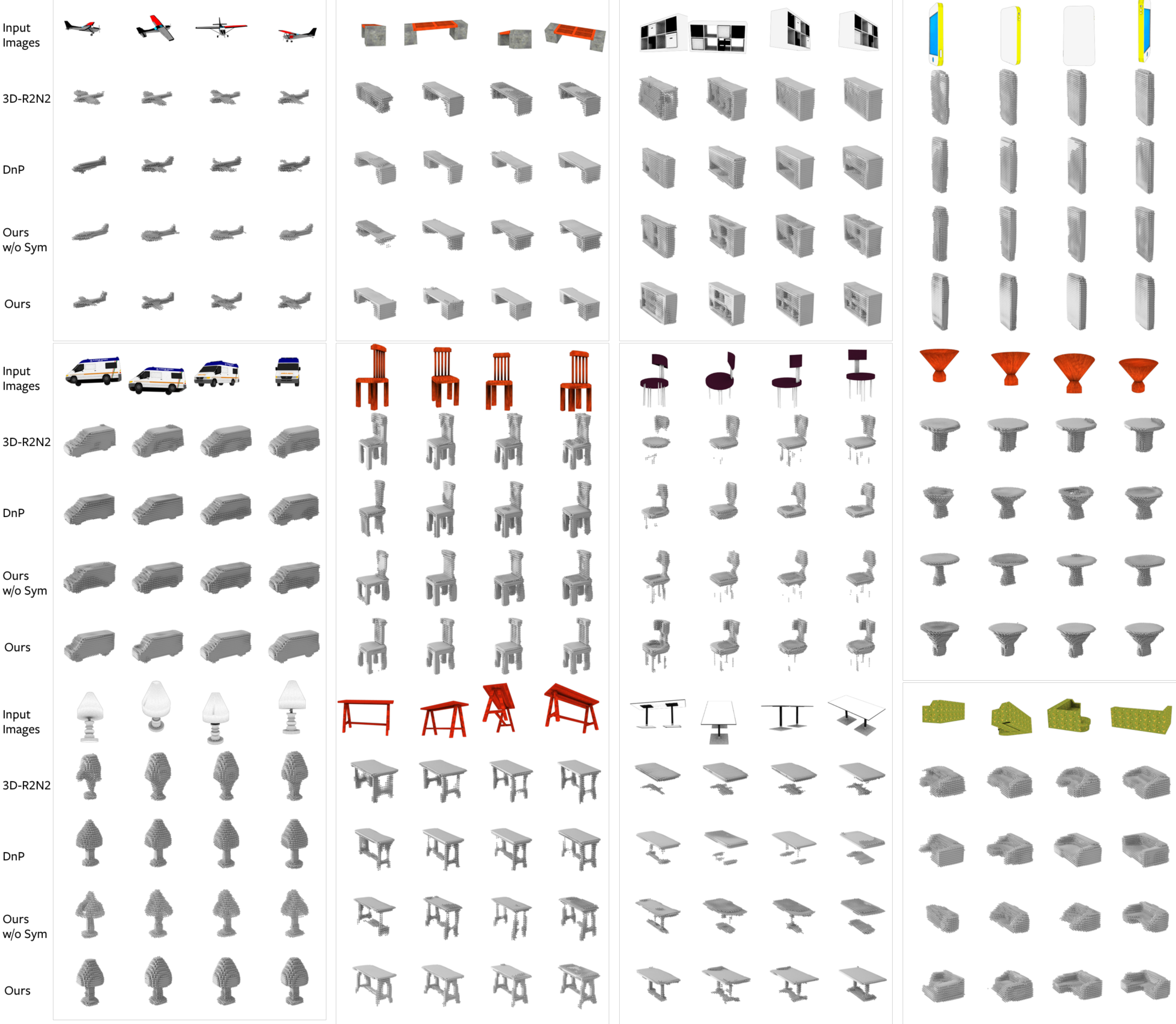}
\caption{\label{fig:vol_pred} {\bf Volumetric prediction}. We show results for 8 example objects. For each example, the top row shows the 4 input views. The second to fifth rows show reconstructed shapes from different methods. The columns correspond to results when using the initial 1, 2, 3, or all 4 views as input.}
\end{figure*}

\begin{figure*}[t]
\center
\includegraphics[width=1.02\linewidth]{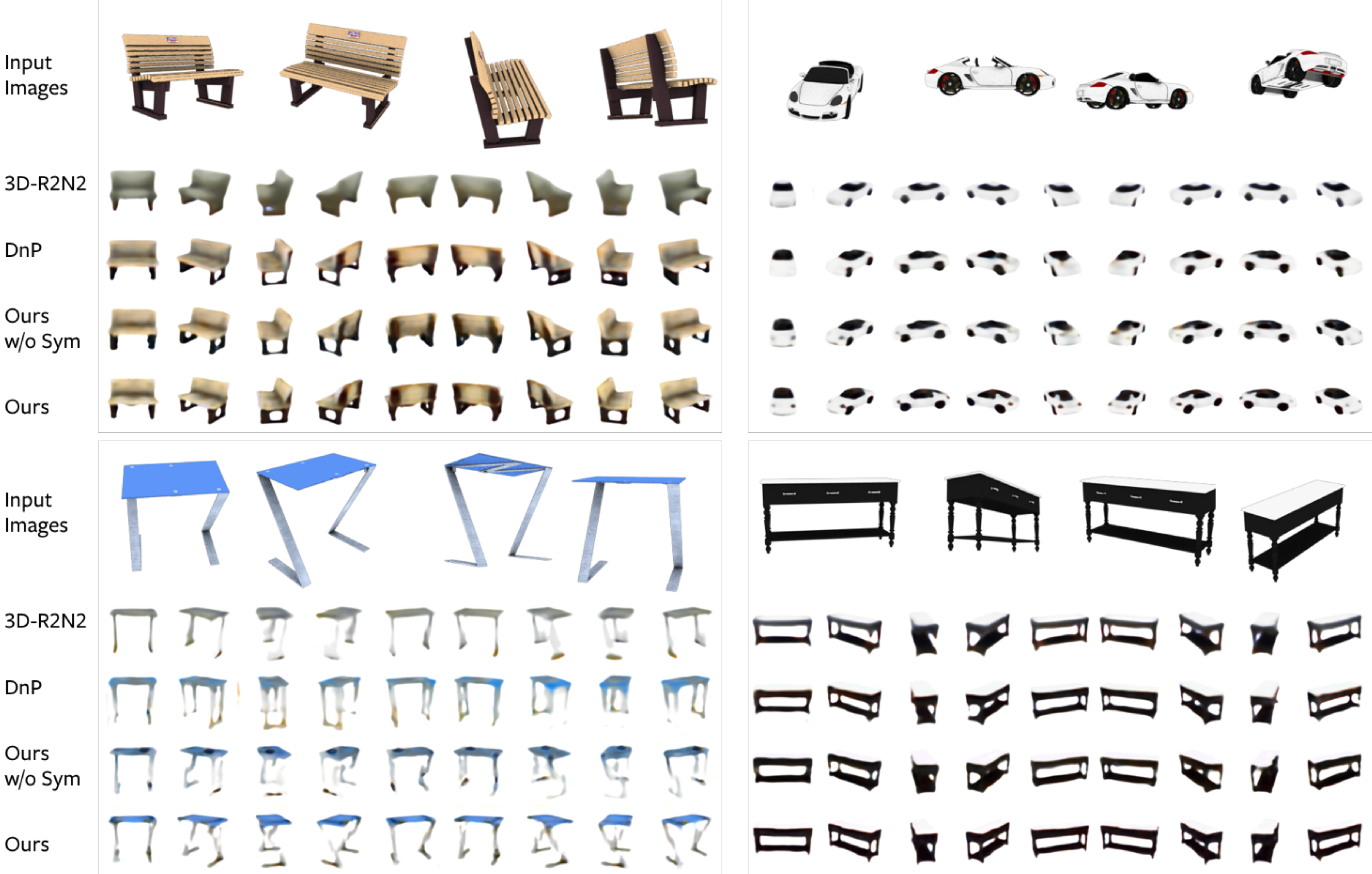}
\caption{\label{fig:view_pred} {\bf Novel view synthesis}. We show novel views synthesized for 4 example objects. The views are generated through a neural renderer from the aggregate representation. Best viewed in color with zoom in.}

\end{figure*}

\section{Experiments}
\subsection{Training Setup}
\paragraph{Dataset.}
We use the ShapeNet dataset~\cite{chang2015shapenet} for empirical validation of our approach. We use models from 13 object categories (similar to previous approaches ~\cite{choy20163d,lsmKarHM2017}), and use random train/val/test splits with (0.7, 0.1, 0.2) fraction of the models. We render each instance from 10 randomly sampled camera viewpoints with azimuth $\in [0, 360)$, elevation $\in [-20, 40]$ degrees, and an additional random camera translation $\in [-0.1,0.1]$ units in each dimension (where ShapeNet models lie in a unit diameter ball). We train our model (and all baselines) using $K=4$ input images during training, and use $K'=5$ images as supervision for learning novel view synthesis. We use voxelized representations of the models with a grid size $32$ for training and evaluating the volumetric reconstruction. 

\paragraph{Baselines.}
We compare our method with several approaches that aggregate multiple images:

\textit{a) 3D-R2N2}: an implicit LSTM based aggregation. To enable view synthesis we extend the model originally presented in~\cite{choy20163d} with a learned decoder to upsample the hidden state to a spatial resolution (similar to our representation $V$), followed by a renderer as described in \secref{tasks3d}.

\textit{b) Depth and Pose based Aggregation (DnP)}: Instead of our symmetry-aware object-centric coordinates from 2D images, one can predict a camera-centric per-pixel depth and the global camera pose. We therefore present a baseline which replaces our canonical coordinate prediction in \secref{lifting} with depth and pose (trained with corresponding supervision), while keeping all other aspects unchanged.

\textit{c) Ours (w/o symmetry)}: To highlight the importance of allowing possible symmetries, we present a baseline which does not leverage symmetry, and instead predicts a single coordinate per pixel.

\begin{figure}[t!]
\minipage{0.49\textwidth}
\subfloat[\label{fig:vol_plot}]{%
  \includegraphics[width=0.48\textwidth]{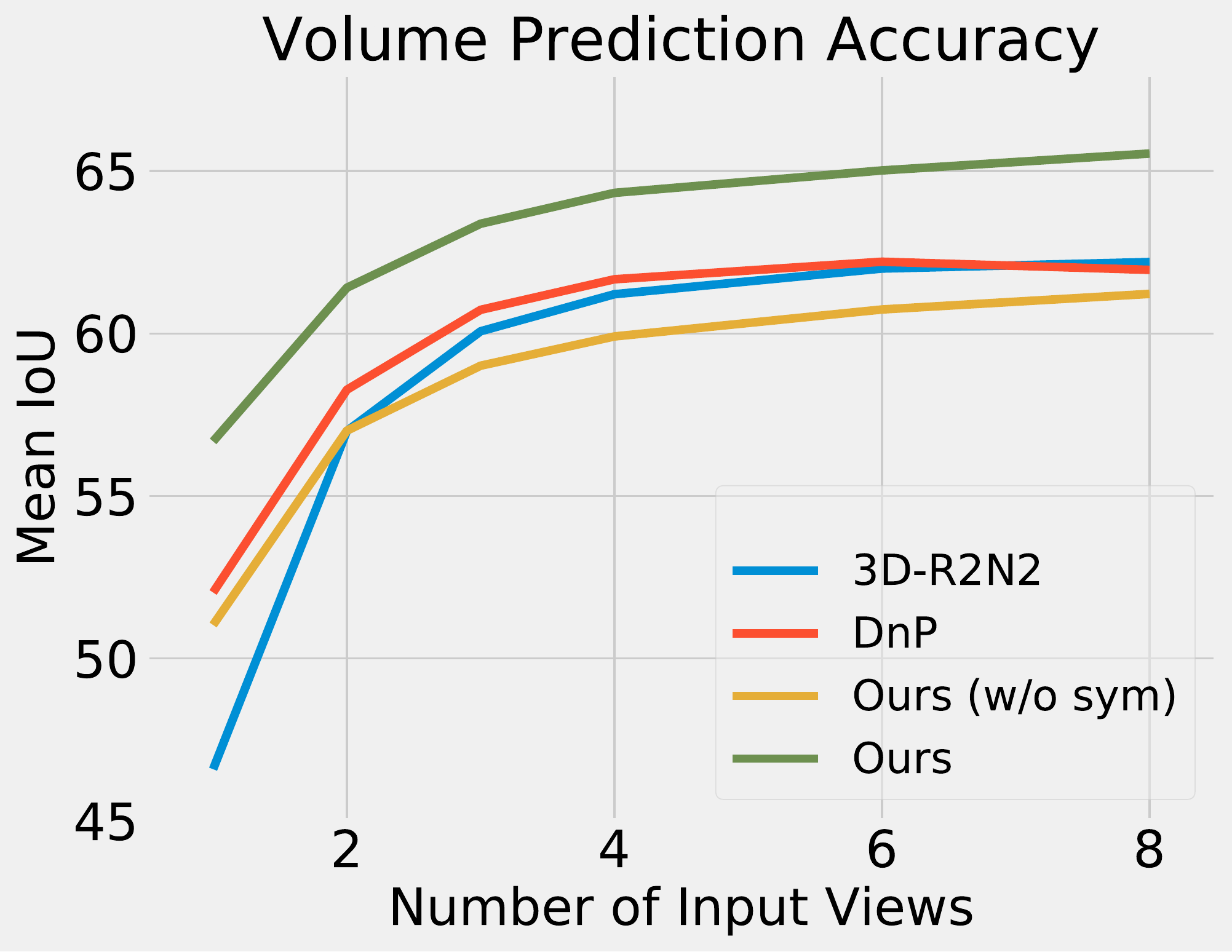}}
  \hfill
  \subfloat[\label{fig:view_plot}]{%
  \includegraphics[width=0.48\textwidth]{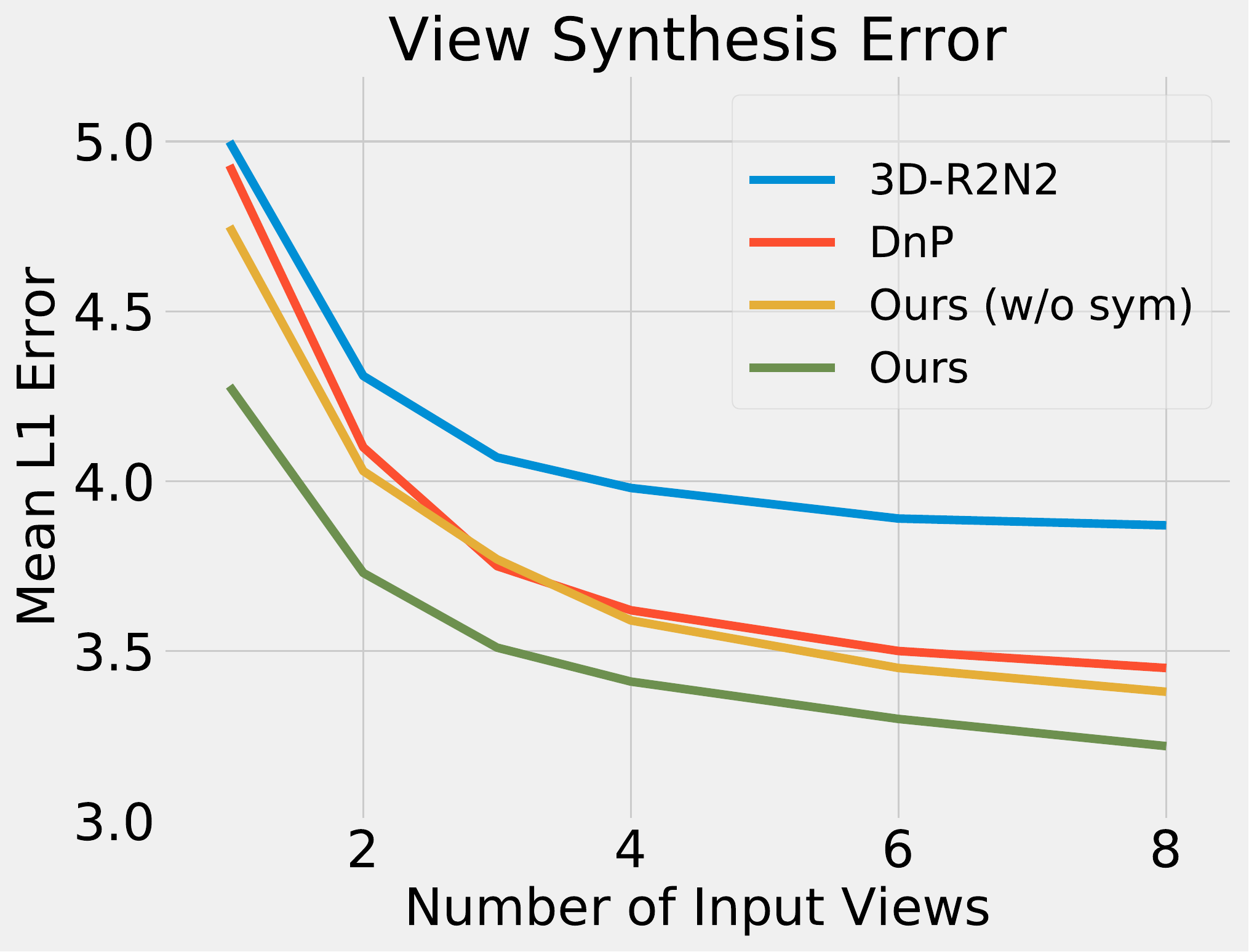}}

\caption{(a) Mean voxel IoU and (b) mean image L1 error vs number of input views.}
\endminipage\hfill
\minipage{0.49\textwidth}
  \includegraphics[width=\linewidth]{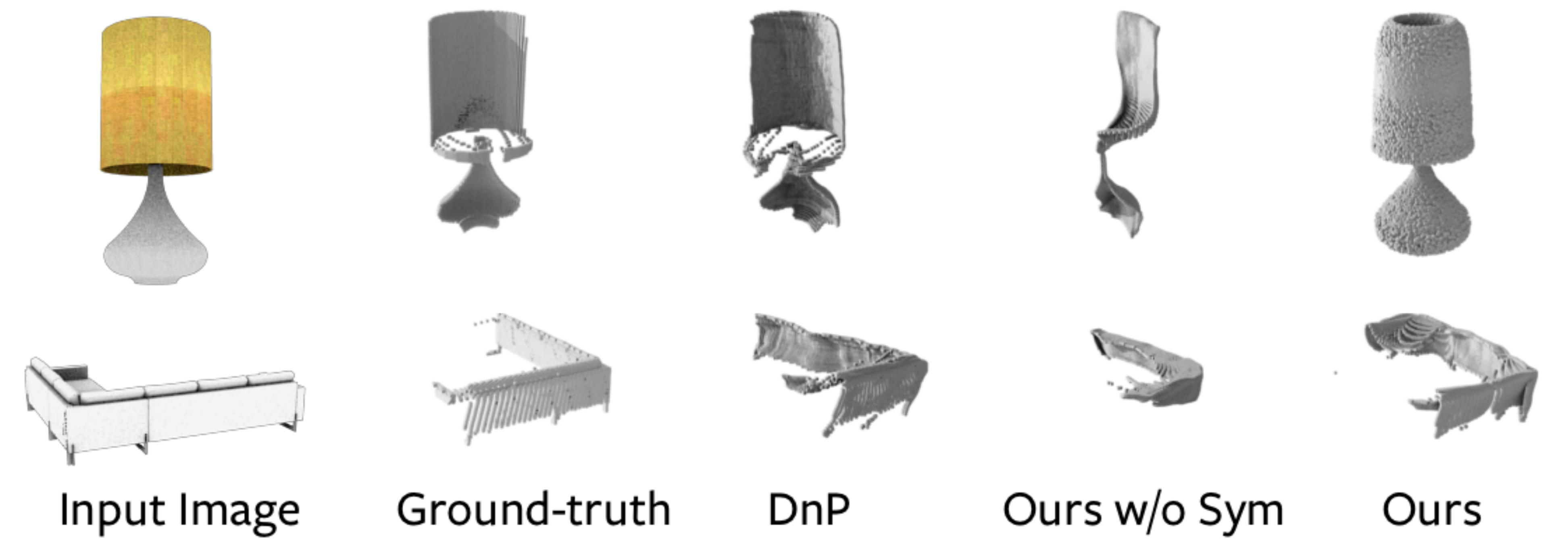}
  \caption{Predicted object-centric coordinates (visualized as point clouds) from a {\bf{single}} image.}\label{fig:nocs_pred}
\endminipage
\end{figure}

\subsection{Evaluation}
\paragraph{Volumetric Reconstruction.} We measure the performance of methods using the intersection over union (IoU) between the predicted and ground-truth $32^3$ volumes. We report the mean IoU score across the 13 categories. As all approaches predict a continuous probability, we report performance for each method using the corresponding optimal binarization threshold (typically around $0.4$). The performance of various approaches in the setting with 4 input views is reported in \tableref{mv3d4v}.

\input{tables/vol_results.tex}
\input{tables/render_results.tex}
We consistently improve over the alternatives of leveraging implicit aggregation or camera-centric prediction. We also clearly see the benefits of incorporating symmetry, in particular for classes such as bench, lamp, and table. While all methods were trained with exactly 4 input views, we also test their performance with fewer/more views at inference and visualize the mean IoU in Fig \ref{fig:vol_plot}. Our approach outperforms the baselines over the spectrum, and performance consistently increases with additional views. Although all methods were similarly trained using 4 input views, we observe a more significant improvement over baselines when using smaller number of input views, indicating that the symmetries allow us to better leverage the information. We visualize in Figure \ref{fig:vol_pred} the predictions with varying number of input views.

\paragraph{Novel-view Synthesis.} We evaluate the performance for the task of view synthesis using L1 error between predicted and ground-truth images (using $5$ novel views per instance). We report the category-wise mean error for various approaches in the setting with 4 input views in \tableref{mvrender}, and highlight  the mean error across classes with varying input views in Fig \ref{fig:view_plot}. We also visualize sample results in the setting with 4 input views in Figure \ref{fig:view_pred}.

We notice a similar trend as in the case of volumetric prediction -- our method improves over the baselines, and error reduces with additional views. In particular, the implicit aggregation method~\cite{choy20163d} has a spatially low resolution aggregate feature, which prevents it from capturing appearance details well. While we note that all approaches produce slightly blurry results, and could be improved with additions \eg adversarial losses, better renderer \etc, our goal here is to highlight the benefits our our aggregation method in comparison to alternates.

\begin{figure}[t]
\center
\begin{overpic}[width=\linewidth,trim=0 0 0 0]{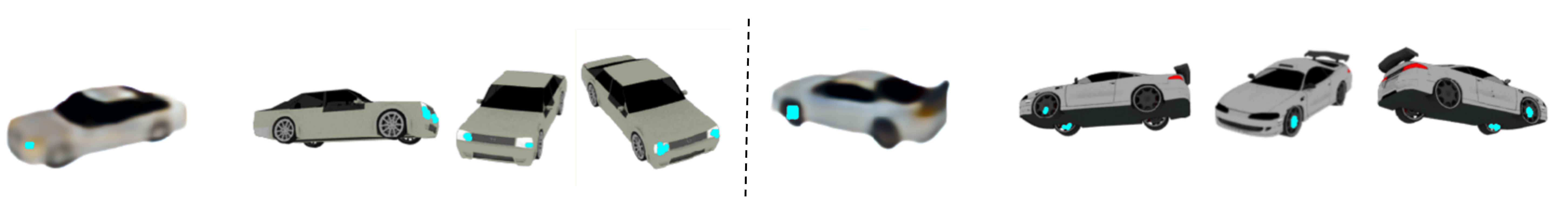}
\put(0,13){\footnotesize Rendered view}
\put(25,13){\footnotesize Input views}
\put(49,13){\footnotesize Rendered view}
\put(75,13){\footnotesize Input views}
\end{overpic}
\caption{\label{fig:grad_vis} Back-propagating the gradient from a region (painted blue) in the rendered view (left) to the original images (right) reveals correspondence and symmetry.}
\end{figure}

\begin{figure}[t!]
    \centering
    \includegraphics[width=\linewidth]{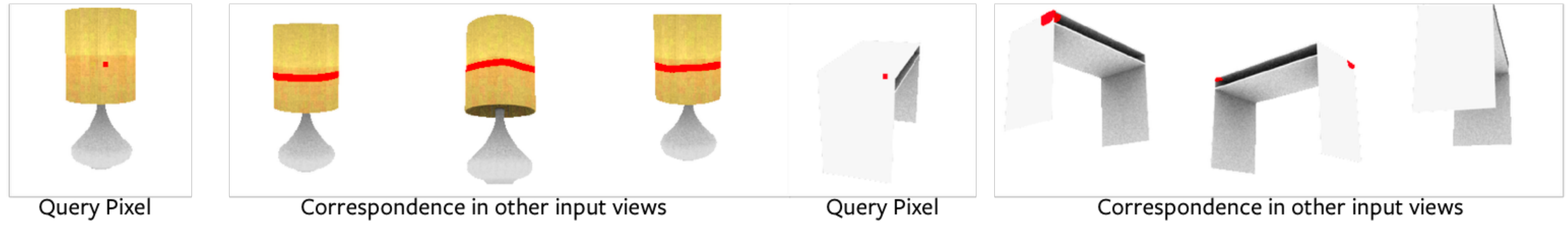}
    \caption{Given a pixel in an input view (left, highlighted in red), we visualize the pixels with the most similar symmetry-aware object coordinates (also highlighted in red).}
    \label{fig:nocs_correspondence}
\end{figure}

\begin{figure}[t!]
    \centering
    \includegraphics[width=\linewidth]{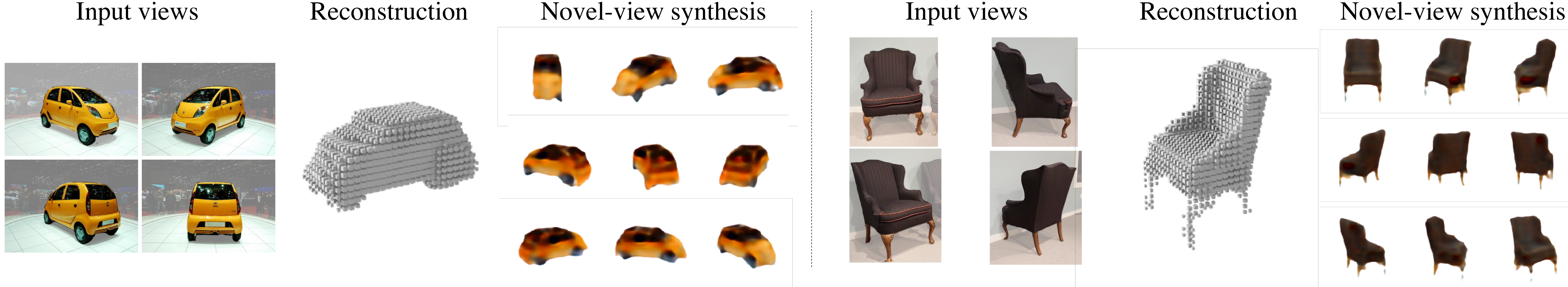}
    \caption{Multi-view aggregation on real images. Given 4 real images of an object in different poses (with object masks), our network is able to reconstruct the object shape and synthesize novel views.}
    \label{fig:real_images}
\end{figure}

\paragraph{Representation Analysis.}
We inspect the structure learned by our model by back-tracing a pixel in a rendered image to its source set of input views. Specifically, we choose a region in the rendered image and compute the gradient of its sum with respect to the input images. We then highlight the source pixels with the most significant gradient magnitude. The result for several such regions is shown in Figure \ref{fig:grad_vis}. We observe that the network relies on the information at the corresponding (or symmetric) pixels in the input images to render the target image \eg the pixels on \textit{both} front wheels in the source images are most influential for rendering the front \textit{left} wheel in a novel views.

We also visualize object-centric coordinates obtained from a single image in Figure \ref{fig:nocs_pred} (by selecting for each pixel the most likely predicted symmetry type), and compare our predictions to those obtained without symmetry inference, or  via predicted per-pixel depth and global camera pose. We notice that our predictions are better aligned, and predict additional points for a symmetric object.

We additionally visualize in Figure \ref{fig:nocs_correspondence} the corresponding pixels in other views that have similar symmetry-aware coordinate prediction to a given query pixel, and observe that the correspondences do respect the symmetry.

\paragraph{Qualitative Results on Real Images.} There are several real-world scenarios that correspond to our inference setups, namely multiple images of an object without access to camera pose. These include for example an object on display on a turntable, images of products online \etc We show some qualitative results of our learned network on such data in Figure \ref{fig:real_images} using segmented images of a rotating car, and chair images from an online seller.

%% file: tables/vol_results.tex
\renewcommand{\arraystretch}{1.4}
\setlength{\tabcolsep}{5pt}
\begin{table*}[t!]
\centering
\vspace{0pt}
\centering
\footnotesize
\resizebox{0.98\linewidth}{!}{
\begin{tabular}{lccccccccccccc|c}
    \toprule
    \textbf{Classes}     &  \textbf{aero}     &  \textbf{bench}     &  \textbf{cabinet}     &  \textbf{car} &    \textbf{chair}     &  \textbf{display}     &  \textbf{lamp}     &  \textbf{speaker} & \textbf{rifle}     &  \textbf{sofa}     &  \textbf{table}     &  \textbf{phone} & \textbf{vessel} & \textbf{mean}\\    
    \midrule
    \textbf{3D-R2N2}~\cite{choy20163d}  &  58.0 & 55.0 & 71.5 & 79.3 & 57.2 & 51.4 & 46.4 & 65.2 & 61.4 & 67.7 & 60.0 & 70.4 & 59.5 & 61.8  \\
    \textbf{DnP} & 57.6 & 56.1 & 70.7 & 80.1 & 61.8 & \bf{54.1} & 46.4 & 66.2 & 65.2 & 70.5 & 60.1 & 69.8 & 58.8 & 62.9  \\    \textbf{Ours (w/o sym)} & 55.8 & 52.0 & 69.5 & 78.7 & 58.2 & 49.5 & 42.3 & 64.1 & 63.5 & 69.2 & 55.4 & 65.8 & 56.5 & 60.0  \\
    \textbf{Ours}  &  \bf{58.6} & \bf{59.4} & \bf{74.0} & \bf{80.3} & \bf{62.1} & 53.0 & \bf{49.0} & \bf{66.6} & \bf{66.0} & \bf{72.6} & \bf{65.0} & \bf{71.6} & \bf{60.6} & \bf{64.5}  \\
    \bottomrule \\
  \end{tabular}}
\caption{\small{Mean voxel IoU for 3D shape reconstruction with 4 input views.}}
\tablelabel{mv3d4v}
\end{table*}

%% file: tables/render_results.tex
\renewcommand{\arraystretch}{1.4}
\setlength{\tabcolsep}{5pt}
\begin{table*}[htb!]
\centering
\vspace{0pt}
\centering
\footnotesize
\resizebox{0.98\linewidth}{!}{
\begin{tabular}{lccccccccccccc|c}
    \toprule
    \textbf{Classes}     &  \textbf{aero}     &  \textbf{bench}     &  \textbf{cabinet}     &  \textbf{car} &    \textbf{chair}     &  \textbf{display}     &  \textbf{lamp}     &  \textbf{speaker} & \textbf{rifle}     &  \textbf{sofa}     &  \textbf{table}     &  \textbf{phone} & \textbf{vessel} & \textbf{mean}\\    
    \midrule
    \textbf{3D-R2N2}~\cite{choy20163d}  &  1.54 & 3.40 & 4.43 & 4.03 & 4.36 & 4.76 & 2.58 & 5.92 & 1.89 & 3.80 & 3.85 & 4.57 & 2.52 & 3.67  \\
    \textbf{DnP} &  1.42 & 3.12 & 4.13 & 3.47 & 3.74 & 4.44 & 2.40 & 5.47 & 1.61 & 3.21 & 3.68 & 3.86 & 2.40 & 3.30  \\    \textbf{Ours (w/o sym)} &  1.44 & 3.30 & 3.77 & 3.43 & 3.86 & 4.40 & 2.63 & 5.44 & 1.65 & 3.15 & 3.94 & \bf{3.79} & 2.44 & 3.33  \\
    \textbf{Ours}  & \bf{1.41} & \bf{2.87} & \bf{3.61} & \bf{3.19} & \bf{3.63} & \bf{4.25} & \bf{2.28} & \bf{5.27} & \bf{1.58} & \bf{2.98} & \bf{3.13} & 4.06 & \bf{2.29} & \bf{3.12}  \\
    \bottomrule \\
  \end{tabular}}
\caption{\textbf{Novel view synthesis.} \small{Mean L1 error (scaled by 100) across classes when using 4 input images for inference.}}
\tablelabel{mvrender}
\end{table*}

%% file: discussion.tex
\section{Discussion}
We have presented an approach for aggregating multiple images of an object instance via predicting symmetry-aware object-centric coordinates, and have demonstrated that this aggregate representation can be leveraged for certain 3D tasks. While this has allowed us to improve over implicit, or camera-centric prediction based aggregation, our approach also has certain shortcomings. Classical SfM methods `lift' pixels to 3D via reasoning across multiple images, as they rely on correspondence across images to do so. Our approach instead does this independently per image, and while the refinement could correct certain errors, the lifting itself could be improved via multi-image reasoning. Further, while the classical reconstruction methods are inherently `unsupervised', our reliance on learning requires the use of supervisory data and it would be a desirable direction to lighten this burden. Lastly, our approach has tackled an object reconstruction setting using normalized coordinates, and it would be interesting to formulate extensions that could handle generic scenes.

%% file: appendix.tex
\section*{Appendix}

\subsection*{A1. Additional Results and Visualization}
We visualize in Figures ~\ref{fig:random-vol-results} and ~\ref{fig:random-render-results} visualizations for volumetric prediction and novel view synthesis for randomly sampled instances. We observe occasional artifacts across methods when rendering empty regions, possibly due to unsuppressed outliers that that contribute features to those regions. We also visualize in Figure \ref{fig:sym_pred} per-symmetry type probability and coordinate predictions, and observe that the network does infer the underlying rotational symmetry. Interestingly, towards centre regions of objects, it tends to infer continuous symmetry as it does not increases the prediction loss.

\input{visuals/random_vol_vis.tex}
\input{visuals/random_render_vis.tex}

\subsection*{A2. Implementation Details}

\paragraph{Feature Splatting.} 
When describing the `lifting' of 2D features to the 3D volumetric grid using their associated coordinates, we evoked a 3D feature grid $\mathcal{V}(x,f)$ corresponding to placing a feature $f$ at coordinate $x$. Intuitively, this operation corresponds to computing the 8 volumetric cells that surround the coordinate $x$, and storing $w_{cell}*f$ at that location, where $w_{cell}$ is the trilinear sampling weight of the cell w.r.t coordinate $x$. Assuming the coordinate of the centre of a cell is $\bar{x}$, and the unit sized feature grid is divided in  $C$ cells along each dimension, the weight $w_{cell}$ is $\prod_{d=1}^3\mathcal{B}(x_d, \bar{x}_d)$, where $\mathcal{B}(x_1, x_2) = \max(0, 1 - C|x_1-x_2|)$.
Note that the weights across cells sum to $1$, and therefore each feature is `spread' across 8 locations. 

\paragraph{Renderer.} As our renderer, we adapt the architecture proposed in the `DeepVoxels' approach by Sitzmann \etal ~\cite{sitzmanndeepvoxels}. We briefly summarize the renderer here, and refer the reader to ~\cite{sitzmanndeepvoxels} for additional details. Given our semi-latent feature representation $V$, which spatially corresponds to an origin-centred unit cube, and a query camera viewpoint $\pi'$, we compute a camera-centric spatial feature grid. To obtain this feature grid, we create a $56~X~56$ spatial grid, and at each pixel, sample features from the 3D volume at $64$ increasing depth values along the ray corresponding to that pixel. Given these per-pixel depth-wise features, a lightweight `occlusion network' predicts a per-pixel probability for using features at each depth, which are then used to average and obtain a single per-pixel feature. This spatial feature grid is then decoded (via a learned decoder) to a rendering of spatial resolution $224~X~224$.

\begin{figure*}[t!]
\center
\includegraphics[width=0.9\linewidth]{./figures_supp/coord_prob_vis.png}
\caption{\label{fig:sym_pred} {\bf Per-symmetry type coordinate and probability predictions}. We visualize the coordinate and probability predictions for the five allowed symmetry types (identity, reflectional symmetry, 2-way, 4-way, and continuous rotational symmetry) On the left we show 4 input images, in the middle the predicted coordinates mapped (colors indicate $xyz$ values), and the probabilities on the right. The five rows correspond to different symmetry types.}
\end{figure*}

\paragraph{Baseline Implementations.}
We train all the baselines on our dataset for 12 epochs using four input views during training. All the baselines use the same renderer as our approach to yield novel views. As the model from Choy \etal~\cite{choy20163d} computes a lower resolution aggregate feature, it is first upsampled via learned up-convolutions to a similar resolution as our feature grid representation, and then passed to the renderer.

The depth and pose based aggregation approach can be considered as an alternate way of computing the coordinates in the normalized space for each pixel. Instead of our approach of directly predicting the coordinates (while accounting for symmetries), this baseline predicts the global camera pose and a per-pixel depth in the camera frame. Given known camera intrinsics (something that our approach does not require for aggregation), these can then be used to compute the 3D coordinate for each pixel. We train this baseline using supervision for both depth and pose prediction, where a 2D UNet~\cite{Ronneberger_2015} outputs per-pixel depth, and a ResNet-18 based CNN (initialized with ImageNet pretraining) learns to predict the 5-dof camera pose -- camera azimuth, elevation, and translation.

We note that previous geometry-driven approaches for multi-view aggregation are not directly applicable to our setting as they crucially rely on ground-truth camera pose for inference. However, the `Depth and Pose' approach is in principle analogous to MVS approaches \eg~\cite{yao2018mvsnet} that infer depth and then aggregate, with the addition of predicting the camera pose.


%% file: visuals/random_vol_vis.tex
\begin{table*}[h!]
\setlength{\tabcolsep}{0.05em}
\centering
  \scalebox{0.85}{
\begin{tabular}{c|c|c|c}
\addpic{figures_supp/id0050_vol_fig.png}  & 
\addpic{figures_supp/id0250_vol_fig.png}  & 
\addpic{figures_supp/id0450_vol_fig.png} &
\addpic{figures_supp/id0750_vol_fig.png}  \\
\midrule
\addpic{figures_supp/id0950_vol_fig.png}  & 
\addpic{figures_supp/id1150_vol_fig.png} &
\addpic{figures_supp/id1350_vol_fig.png} &
\addpic{figures_supp/id1650_vol_fig.png}  \\ 
\midrule
\addpic{figures_supp/id1850_vol_fig.png}  & 
\addpic{figures_supp/id2050_vol_fig.png} &
\addpic{figures_supp/id2250_vol_fig.png} &
\addpic{figures_supp/id2550_vol_fig.png}  \\
\midrule
\addpic{figures_supp/id2750_vol_fig.png}  & 
\addpic{figures_supp/id2950_vol_fig.png} &
\addpic{figures_supp/id3150_vol_fig.png} &
\addpic{figures_supp/id3450_vol_fig.png}  \\ 
\midrule
\addpic{figures_supp/id3650_vol_fig.png}  & 
\addpic{figures_supp/id3850_vol_fig.png} &
\addpic{figures_supp/id4050_vol_fig.png} &
\addpic{figures_supp/id4350_vol_fig.png}  \\ 
\end{tabular}
}
\vspace{-0.1in}
\captionof{figure}{{\bf Volumetric prediction}. We show results for randomly sampled instances. The top row shows the 4 input views, and the second to fifth rows show reconstructed shapes from different methods (3D-R2N2, DnP, Ours w/o Sym, and Ours). The columns correspond to results when using the initial 1, 2, 3, or all 4 views as input.}
\label{fig:random-vol-results}
\end{table*} 

%% file: visuals/random_render_vis.tex
\begin{table*}[h!]
\setlength{\tabcolsep}{0.05em}
\centering
  \scalebox{0.85}{
\begin{tabular}{c|c|c|c}
\addpicwide{figures_supp/id0000_rendering_fig.png}  & 
\addpicwide{figures_supp/id0300_rendering_fig.png}  & 
\addpicwide{figures_supp/id0600_rendering_fig.png}  \\
\midrule
\addpicwide{figures_supp/id0800_rendering_fig.png}  & 
\addpicwide{figures_supp/id1200_rendering_fig.png} &
\addpicwide{figures_supp/id1500_rendering_fig.png}  \\ 
\midrule
\addpicwide{figures_supp/id1800_rendering_fig.png}  & 
\addpicwide{figures_supp/id2100_rendering_fig.png} &
\addpicwide{figures_supp/id2400_rendering_fig.png}  \\
\midrule
\addpicwide{figures_supp/id2700_rendering_fig.png}  & 
\addpicwide{figures_supp/id3000_rendering_fig.png} &
\addpicwide{figures_supp/id3300_rendering_fig.png}  \\ 
\midrule
\addpicwide{figures_supp/id3600_rendering_fig.png}  & 
\addpicwide{figures_supp/id3900_rendering_fig.png} &
\addpicwide{figures_supp/id4200_rendering_fig.png}  \\ 
\end{tabular}
}
\vspace{-0.1in}
\captionof{figure}{{\bf Novel view synthesis}. We show results for randomly sampled instances. The top row shows the 4 input views, and the second to fifth rows show reconstructed shapes from different methods (3D-R2N2, DnP, Ours w/o Sym, and Ours) from 9 novel views.}
\label{fig:random-render-results}
\end{table*} 